\documentclass[10pt,twocolumn,letterpaper]{article}

\usepackage{wacv}
\usepackage{times}
\usepackage{epsfig}
\usepackage{graphicx}
\usepackage{amsmath}
\usepackage{amssymb}
\usepackage{booktabs}

\def\wacvPaperID{1006} 

\wacvapplicationstrack 

\wacvfinalcopy 

\ifwacvfinal
\usepackage[breaklinks=true,bookmarks=false]{hyperref}
\else
\usepackage[pagebackref=true,breaklinks=true,colorlinks,bookmarks=false]{hyperref}
\fi

\newcommand{\myheading}[1]{\vspace{1ex}\noindent \textbf{#1}}
\newcommand{\Sref}[1]{Sec.~\ref{#1}}

\newcommand{\Fref}[1]{Fig.~\ref{#1}}
\newcommand{\Tref}[1]{Table~\ref{#1}}
\def\mL{\mathcal{L}}

\begin{document}

\title{Driver Attention Tracking and Analysis}

\author{%
Dat Viet Thanh Nguyen$^{1}$  \quad Anh Tran$^{1}$ \quad  Hoai Nam Vu$^{2}$ \quad
Cuong Pham$^{1,2}$ \quad
Minh Hoai$^{1,3}$  \\
\small{\textsuperscript{1}VinAI Research, Vietnam \quad \textsuperscript{2} Posts \& Telecommunications Inst. of Tech., Vietnam \quad 
\textsuperscript{3}The University of Adelaide, Australia} \\
\texttt{\scriptsize \{v.datnvt2, v.anhtt152\}@vinai.io} \quad \texttt{\scriptsize  namvh@ptit.edu.vn} \quad \texttt{\scriptsize \{v.cuongpv11, v.hoainm\}@vinai.io} 
}


\maketitle
\thispagestyle{empty}

\begin{abstract}
We propose a novel method to estimate a driver's points-of-gaze using a pair of ordinary cameras mounted on the windshield and dashboard of a car. This is a challenging problem due to the dynamics of traffic environments with 3D scenes of unknown depths. This problem is further complicated by the volatile distance between the driver and the camera system. To tackle these challenges, we develop a novel convolutional network that simultaneously analyzes the image of the scene and the image of the driver's face. This network has a camera calibration module that can compute an embedding vector that represents the spatial configuration between the driver and the camera system. This calibration module improves the overall network's performance, which can be jointly trained end to end. 

We also address the lack of annotated data for training and evaluation by introducing a large-scale driving dataset with point-of-gaze annotations. This is an in situ dataset of real driving sessions in an urban city, containing synchronized images of the driving scene as well as the face and gaze of the driver. Experiments on this dataset show that the proposed method outperforms various baseline methods, having the mean prediction error of 29.69 pixels, which is relatively small compared to the $1280{\times}720$ resolution of the scene camera. 
\end{abstract}

\section{Introduction}

The objective of this work is to develop an economical device that can monitor the face and head movement of a driver and determine what parts of the traffic scene the driver is attending to. We aim for a device that can be mass-produced and installed in vehicles to assist drivers, enhancing the driving experience and reducing accidents. For example, it can be used as part of a system to alert a driver when they fail to notice an important traffic sign or a dithering jaywalker on the road. We can also use the device to analyze what attracts or distracts the driver's attention, which would help discover potential dangers and design better road intersections.

One approach to track the driver's points of gaze is to use a pair of eye-tracking glasses \cite{Holmqvist_book11}. But eye-tracking glasses are expensive and intrusive, so they are not practical for widespread deployment and everyday usage. Furthermore, eye-tracking glasses are not suitable for our purposes because they do not always tell us what the driver misses, even though they always tell us what the driver looks at. We would not know about the existence of potential danger on the road unless the danger is inside the field of view of the eye-tracking glasses. This is, however, not guaranteed because the field of view of the eye-tracking glasses depends on the direction of the driver's head, and the driver can be distracted by things inside the vehicle.

Given the above limitations of wearable eye trackers, we propose to develop a dashboard-mounted eye-tracking system instead. This system only requires two ordinary cameras, one pointing at the driver's face and one looking out to the driving scene. Our paper's main focus is to develop a computer vision algorithm that takes as input the two synchronized image frames from these cameras and outputs the location of the driver's fixation.

It is, however, challenging to develop such an algorithm. First, there is no publicly available in situ dataset with the right type of image pairs and gaze annotation to train this algorithm. This is perhaps due to the difficulty of determining and annotating the gaze points of a driver in image frames from a camera mounted far away from the driver's eyes. To this end, one contribution of our paper is the collection of a large-scale dataset with more than one hundred thousand image pairs with point-of-gaze annotation. Second, being a dashboard-mounted system, the relative location between the driver and the eye tracker can be changed even within the same driving session. Unfortunately, it would be impractical to ask the driver to calibrate the system while driving or stop driving to calibrate the system. Thus it is important to have an algorithm that can continually adapt and calibrate itself. To this end, the technical contribution of this paper is a self-calibrating gaze estimation method. Specifically, we propose a novel convolutional neural network architecture for estimating the point-of-gaze based on a pair of face and scene images. We treat the relative position between the cameras and the driver as latent variables and use a camera calibration module to estimate their relationship. This camera calibration module is trained together with the entire network in an end-to-end manner. 
Experiments on our dataset show that the camera calibration module significantly improves accuracy of the predicted gaze points. Our method achieve mean prediction error of 29.69 pixels, which is relatively small compared to the $1280{\times}720$ resolution of the scene image. 


\section{Related Work}
In this paper, we develop a method to estimate the driver's gaze-point, and we also introduce a new dataset for this task. In this section, we will review related works in gaze estimation and driver monitoring, and describe the uniqueness of our method and dataset. Note that our objective is to determine the specific point where a person is looking, a goal distinct from the challenges of saliency and scanpath prediction, e.g.,~\cite{m_Wei-etal-NIPS16,m_Zelinsky-etal-NBDT20, m_Yang-etal-CVPR20, m_Chen-etal-SR21, m_Yang-etal-ECCV22, m_Mondal-etal-CVPR23}. These latter tasks focus on forecasting where a person is likely to look.

\myheading{Gaze estimation datasets.} There have been several datasets collected for human gaze estimation in recent years. A well-known example is Eye Chimera \cite{florea2013can}, which consists of 1172 RGB frontal face images manually marked with seven gaze directions during data collection. In the Columbia dataset~\cite{smith2013gaze}, five different cameras were used to capture 880 samples of 56 subjects. The UT Multiview dataset \cite{sugano2014learning} was constructed to investigate appearance-based gaze estimation; it consists of 50 different subjects with 160 gaze directions acquired by eight cameras. The EYEDIAP dataset \cite{funes2014eyediap} contains images captured by two cameras, and the recording method was designed to systematically cover most of the variables that affect the remote gaze estimation algorithms. Gazefollow~\cite{recasens2016they} is a dataset with annotated people and gaze points in the same images. 
VideoAttentionTarget \cite{chong2020detecting} is similar to Gazefollow, but for video frames.  
Gaze360 \cite{kellnhofer2019gaze360} is one of the largest datasets for 3D gaze estimation, which consists of 238 subjects in both indoor and outdoor environments. This dataset was captured across a wide range of head poses, distances, and gaze directions. GazeCapture \cite{krafka2016eye} is a large-scale dataset consisting of 2.5 million image frames from 1450 people, annotated with the points of gaze on mobile devices. WebGazer~\cite{papoutsaki2018eye} is a webcam dataset of 51 subjects, recording the points of gaze of web users browsing the Internet. Unfortunately, none of the existing datasets can be used for our task because they either do not have point-of-gaze annotation (only gaze direction) or only have point-of-gaze information on digital screens. None of these datasets contain point-of-gaze information of drivers on 3D street scenes.





\myheading{Driver monitoring datasets.} For monitoring a driver's behavior and attention, several datasets have been collected, which can be divided into three broad categories: hand-based, body-based, and face-based. The hand-based datasets used an ego-centric camera to capture the actions of the drivers' hands \cite{ohn2013power, das2015performance}. The body-based datasets used a side-view camera to capture the upper bodies of drivers. One of the largest datasets of this kind is StateFarm, which contains nine distracted action classes of the driver. The AUC Distracted Driver Dataset \cite{abouelnaga2017real} was collected using the rear camera of an ASUS ZenPhone, capturing the driving behavior of 31 participants from seven different countries. The dataset has more than 17K images annotated according to 10 postures. For better body posture estimation, RGB-D cameras have also been used \cite{craye2015driver,borghi2017poseidon}.

Where a driver is looking at can be estimated based on the his head pose~\cite{sikander2018driver}, and several datasets have been collected for this task. DADA \cite{fang2019dada} and BDD-A \cite{xia2017training} were constructed by defining various fixation zones inside a car such as windshield, center stack, rear-view mirror, and speedometer \cite{vora2018driver, martin2018dynamics}. Gaze information in these datasets, however, has very low spatial resolution due to the limited number of fixation zones. Some datasets were not obtained in real driving environments but captured in a laboratory environment with a driving simulator, where the participants were asked to look at different zones during the data collecting process~\cite{zhang2017mpiigaze, schwarz2017driveahead, roth2019dd}. Similar to our dataset, the DR(eye)VE \cite{palazzi2018predicting} dataset also used smart glasses for collecting the driver's points of gaze. However, it did not use a camera for capturing driver's face and head pose, so cannot be used to directly predict the gaze point. Unlike these datasets, ours was captured in real driving environments. Furthermore, while most existing datasets were captured with only one or two cameras, ours contains synchronized videos from face, scene, gaze cameras. It is a rich dataset for multiple tasks, not just points-of-gaze estimation. 



\myheading{Gaze estimation methods.} There are two types of gaze estimation methods: model-based and appearance-based. The former utilizes geometric characteristics of the human eye to predict gaze direction, such as corneal infrared reflection, iris contour, and pupil center \cite{valenti2008accurate, venkateswarlu2003eye, guestrin2008remote}. One disadvantage of model-based methods is that they require dedicated hardware such as HD cameras, RGB-D cameras, and infrared light sources \cite{alberto2014geometric, xiong2014eye}. Moreover, model-based methods only achieve good accuracy under a short distance between the camera and the human eye, so they are more suitable to use in a laboratory environment with controlled conditions. The appearance-based methods are less restrictive, and they receive increasing attention from the research community. These methods use one or multiple cameras to capture a human face. After that, a mapping function is learned to predict gaze direction from eye images. Different types of mapping functions, including neural network, adaptive linear regression, and Gaussian regression, have been used. More recent methods such as \cite{zhang2015appearance} and \cite{zhu2017monocular} used deep CNNs \cite{LeCun-et-al-NC89} to map a human face image to a gaze direction. In \cite{cheng2020gaze, ali2020deep}, the left and right eye images are fit to different CNN streams, and the output feature maps are combined to obtain the final gaze direction after a fully connected layer. Recently, \cite{kim2020gaze} proposed to use GANs \cite{Goodfellow-etal-NIPS14} to enhance human eye images captured under low-light environments before fitting into a CNN model. \cite{recasens2016they, chong2018connecting, lian2018believe, chong2020detecting, kellnhofer2019gaze360, m_Miao-etal-WACV23} take the advantage of deep network to follow the gaze of the person inside the input image (not the viewer of the image), which are different from ours. Inspired by the success of recent CNN-based models for gaze estimation, we also develop a CNN-based method in this paper. Our method addresses a novel task of estimating the points of gaze on 3D scenes with unknown depths; this is different from existing methods for gaze localization on a 2D display.



\section{Drivers' Points-of-Gaze Dataset \label{sec:dataset}}

We aim to develop a neural network to estimate the points-of-gaze of a driver given a pair of face and scene images. Unfortunately, there was no existing in situ dataset that can be used for training and evaluating this network, so we collected ourselves a dataset called Drivers' Points-of-Gaze (DPoG). The DPoG dataset contains gaze behavioral data of 11 drivers as they drove through the busy streets of an urban city. There were a total of 19 driving sessions, reflecting the real driving conditions that most drivers in this city experience every day. 

\subsection{Data collection and annotation}

\def\subFigSz{0.32\linewidth} 

\begin{figure}
\centering 
\includegraphics[width=\linewidth]{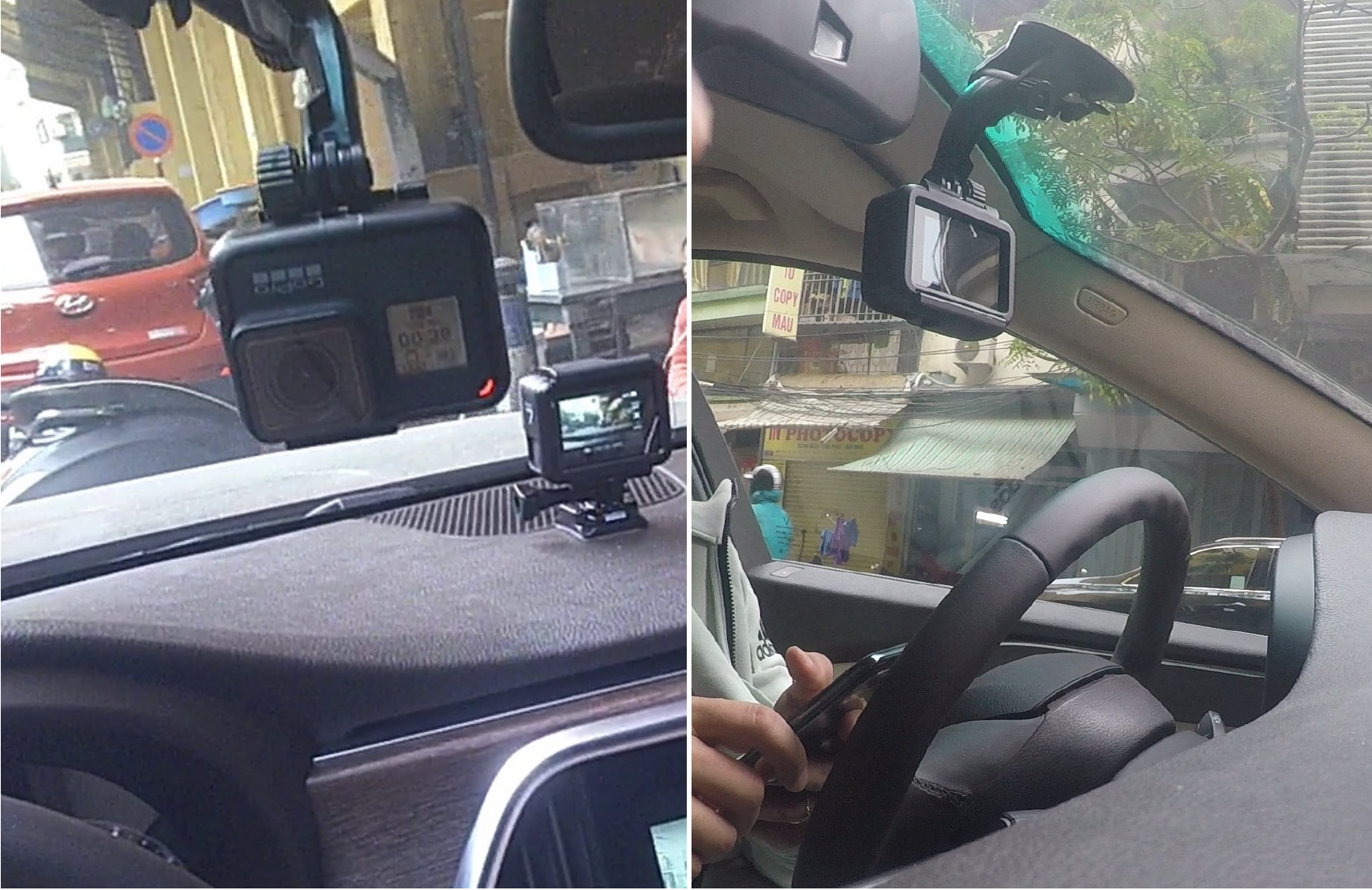}
   \caption{{\bf Positions of GoPro cameras used for data collection.} A  camera was attached to the windshield to capture a driver's face and head movements. Another camera was placed on the dashboard, pointing out to the road. \label{fig:setup}} 
 \vspace{-3mm}
\end{figure}

\myheading{Hardware and setup.} The main hardware components of our system were two GoPro cameras, as shown in Fig. \ref{fig:setup}. One camera was mounted on the windshield, pointing at a driver's face and recording the face and head movements. The second camera was mounted on the dashboard, pointing to the head space of the vehicle. Each GoPro had an on-board SD card, where the recorded videos were stored.
To obtain the ground truth points-of-gaze of the drivers, we used a pair of SMI eye-tracking glasses (Model 2). Note that the eye-tracking glasses were only needed to collect training data; they will not be needed in the final system.

We will refer to the videos captured by the GoPro cameras as face video and scene video, and the video captured by the eye-tracking glasses as gaze video. Face and scene videos were captured at $1280{\times}720$ resolution and 30 frames per second (fps), while gaze video was captured at  $1280{\times}960$ resolution and 24 fps. 

\myheading{Driving sessions.} We recruited a total of 13 drivers (12 males, age range from 23 to 50, normal or corrected-to-normal vision). The data was collected over two weeks at different times of the day. Each driver was asked to participate in two to three driving sessions with a break in between sessions. For each driving session, the driver was asked to drive to a particular destination following a planned route. The duration of a driving session depends on the route and the traffic condition; the minimum, maximum, and mean were 15, 35, and 29.5 minutes respectively. After data collection, we found that the data from several sessions was unusable, being either incomplete (e.g., no data from one of the three cameras) or corrupted (e.g., due to incorrect gaze information). The bad data sessions were subsequently excluded from our dataset. In the end, we had 19 usable driving sessions from 11 drivers. 

\myheading{Calibration.} The eye-tracking glasses were calibrated with a three-point calibration procedure at the beginning of each driving session. The GoPro cameras were purposefully not calibrated, given the fragile relationship between the cameras' positions and the driver's location. 

\myheading{Synchronization.} For synchronization, we asked drivers to clap their hands before each driving session. These clapping hand moments together with audible speech and traffic noise were used to synchronize the GoPro videos. The GoPro videos had a consistent frame rate (30 fps), and it was sufficient to synchronize the two GoPro videos with a single time shift parameter. We computed this time shift parameter so that the shifted audio signals were maximally correlated. 

It was more difficult to synchronize the gaze video with the other two videos. Although the gaze video was shown to be encoded at \mbox{24 fps}, we found that the actual frame rate varied within each video, perhaps due to the quality of the internal clock on board the compact wearable device. Due to this inconsistency issue, it was impossible to use a global parametric model to account for the time lag and frame rate differences between the gaze video and the other two videos. To overcome this problem, we extracted short 30--60 second clips from the gaze video and synchronized each clip individually. We visually inspected and found the corresponding face and scene clips for each extracted gaze clip. Altogether, we collected 589 sets of three synchronized video clips from the three cameras.

\def\subFigSzTwo{0.32\textwidth} 
\begin{figure*}[t]
\begin{center}
\makebox[0.21\textwidth]{\scriptsize{Gaze frame}}
\makebox[0.38\textwidth]{\scriptsize{Warped gaze frame}}
\makebox[0.31\textwidth]{{\scriptsize Scene frame \& transferred gaze point}}
\includegraphics[width=\textwidth]{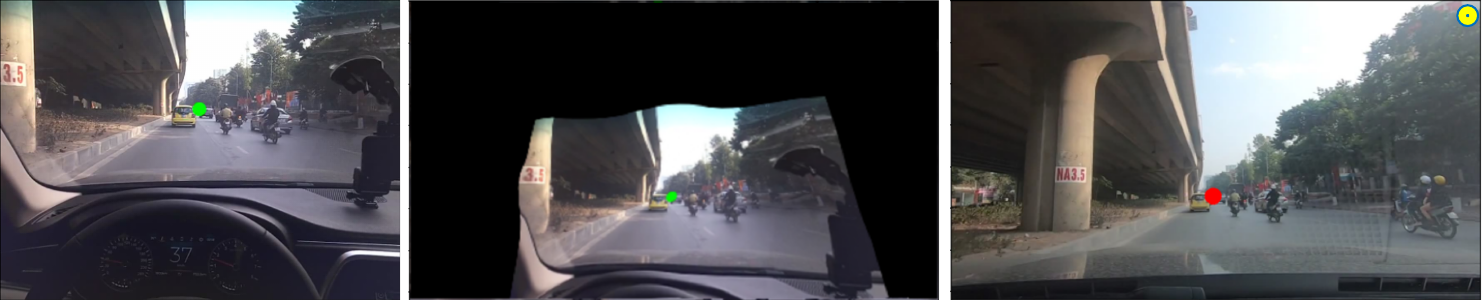}
   \caption{{\bf Matching result using RANSAC-Flow \cite{shen2020ransac}}. RANSAC-Flow is used to warp the gaze frame to the scene frame and transfer the gaze point (green dot) from the gaze frame to the scene frame (red dot). On an annotated dataset of 589 instances, the median and mean errors are 9.2 and 25.1 pixels, which are relatively small compared to the $1280{\times}720$ size of the scene frame. The top right corner of the scene frame shows a circle with the radius of 25.1 pixels.  \label{fig:ransac}}
\vspace{-5mm}
\end{center}
\end{figure*}

\myheading{Point-of-gaze annotation.} From the set of synchronized video clips, we extracted 176,451 triplets of synchronized (scene, face, gaze) frames. Among them, only 152,794 triplets contain gaze information because not every frame from the gaze video contained a gaze point. We used RANSAC-Flow \cite{shen2020ransac} to warp the gaze frame to the scene frame and transfer the gaze point from the gaze frame to the scene frame. Depending on the difference in the perspectives of the gaze and scene frames, RANSAC-Flow might fail and there might not be a corresponding gaze point in the scene frame. We manually verified all the matching results and removed all obviously wrong cases. After this step, only 143,675 frame triplets remained. 
For quality assurance, we sampled one frame triplet randomly from each of 589 short clip triplets and annotated the gaze points in the scene frames manually. On this manually annotated set, the median and mean distances between the transferred gaze point using RANSAC-Flow and the manually annotated gaze point are 9.2 and 25.1 pixels, which are relatively small compared to the scene frame size ($1280{\times}720$ pixels). \Fref{fig:ransac} shows some matching results using RANSAC-Flow.



\myheading{Train and test split.} We divided the data into disjoint training and testing sets, ensuring that the data for each driving session would be used either for training or testing and not for both. After removing bad data sessions due to missing camera view or inaccurate gaze tracking results, we were left with 19 driving sessions from 11 subjects (one subject with three sessions, six subjects with two sessions, and four subjects with one session). For a subject with one session, we put his data into the training set. For a subject with two sessions, we randomly chose one session for training and one for testing. For the subject with three sessions, we randomly chose two sessions for training and one for testing. \Tref{tab:data_stats} displays some statistics of our dataset.



\subsection{Scene and gaze statistics \label{sec:scene_gaze_stats}}

\begin{figure*}[t]
\begin{center}
\includegraphics[width=0.47\linewidth]{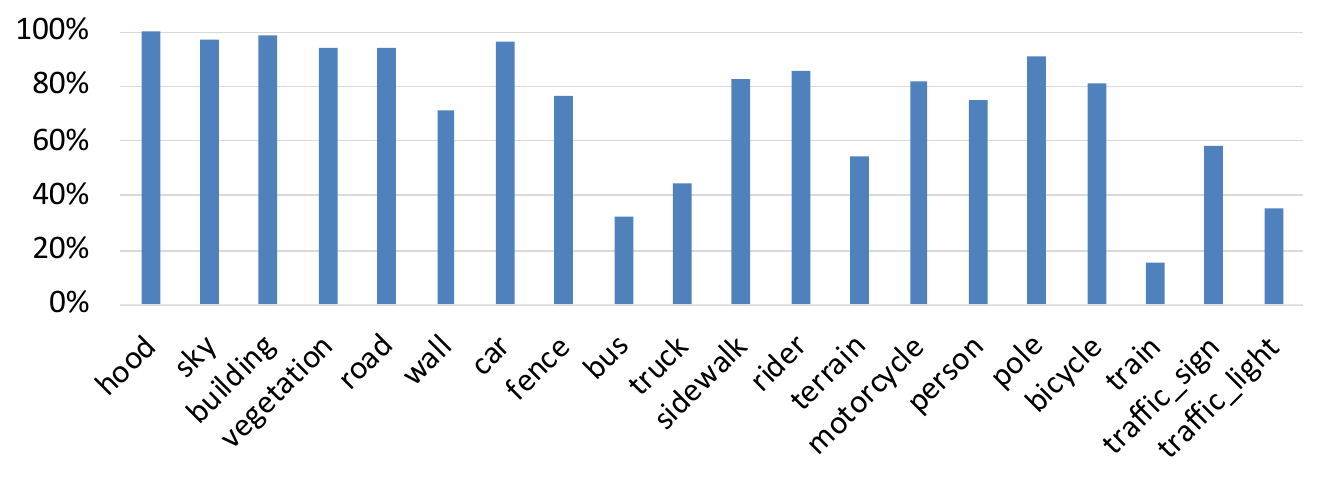} 
\includegraphics[width=0.47\linewidth]{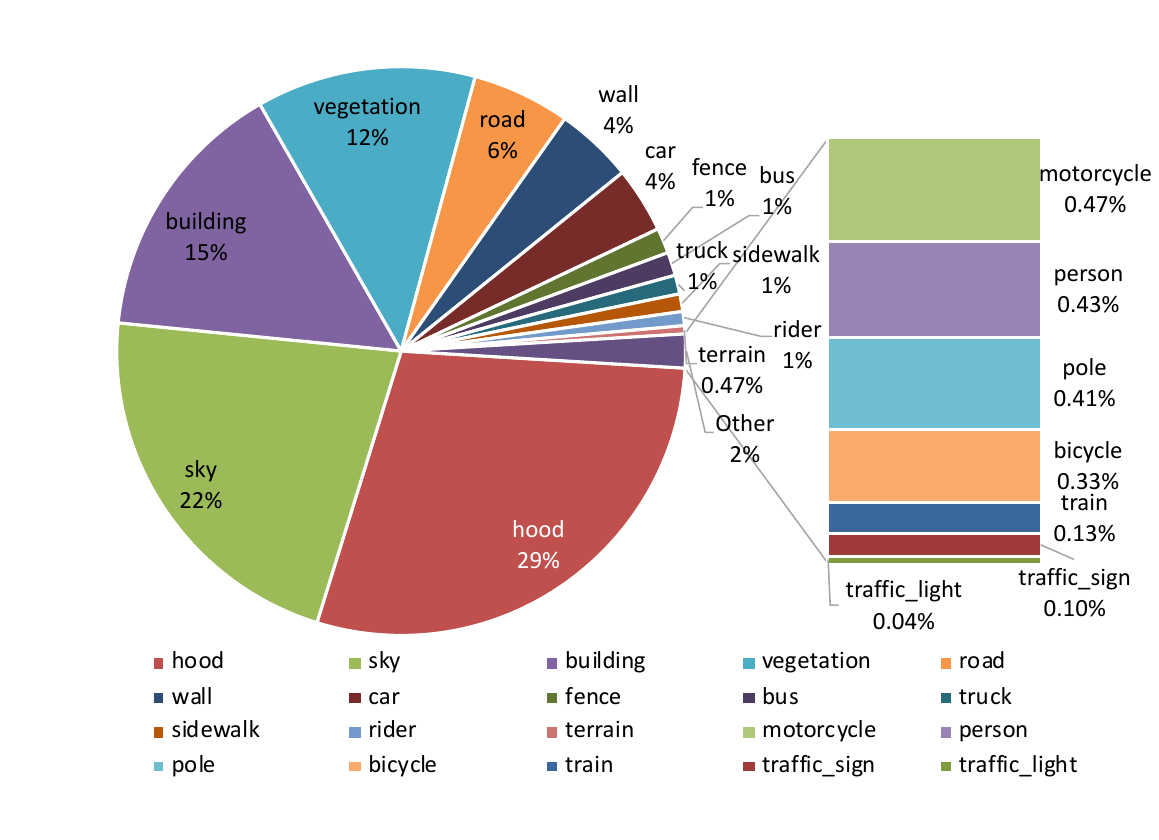}
\makebox[0.47\linewidth]{\small{(a) Scene statistics at image level }}
\makebox[0.47\linewidth]{\small{(b) Scene statistics at pixel level}}
   \caption{{\bf Scene statistics.} (a):  the percentages of images from the scene camera containing objects in each semantic class. Almost all images contain road, sky, building, vegetation, and car. Bicycles and motorcycles are also seen very often. The least appearing classes are train and traffic\_light. (b): the percentages of scene-image pixels belonging to each semantic class. The majority of the pixels belong to road, sky, building, and vegetation.   }
\vspace{-4mm}
\label{fig:scene_stats}
\end{center}
\end{figure*}

\begin{figure}[t]
\centering 
\includegraphics[width=\linewidth]{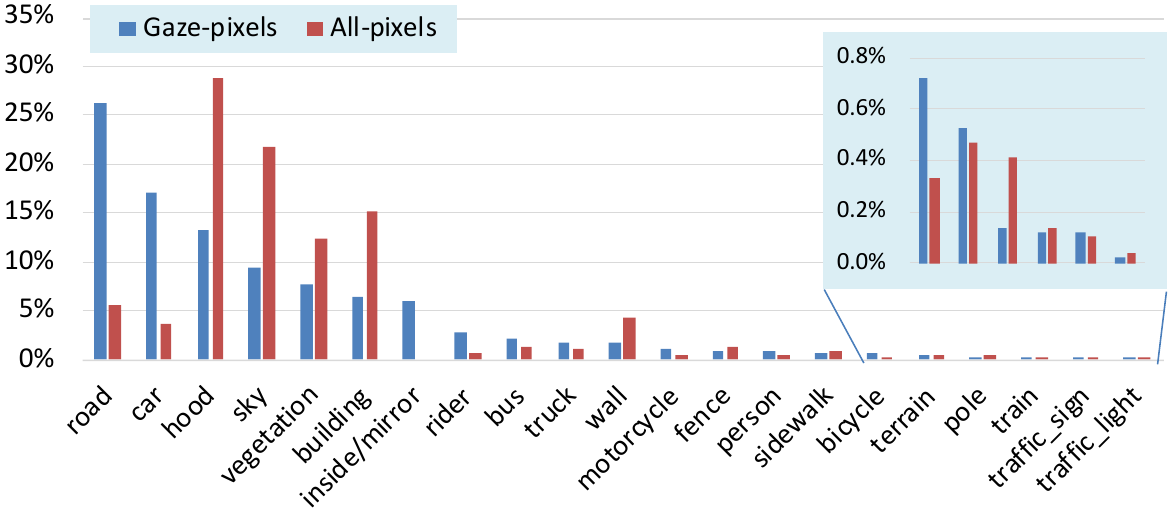}
   \caption{{\bf Semantics of gaze pixels and all pixels.} For each semantic class, the red bar shows the percentage of times a fixation point belongs to the class, and the blue bar is the percentage of pixels in the scene camera belonging to this class. The subplot in the middle is the zoom-in window for the classes with smallest percentages of occurrences. }
\label{fig:gaze_scene_perc}
\vspace{-2mm}
\end{figure}

Figs. \ref{fig:scene_stats} and \ref{fig:gaze_scene_perc} show several scene and behavioral statistics on our data. We used a semantic segmentation method \cite{tao2020hierarchical} to obtain these statistics. The object classes are derived from the cityscapes dataset \cite{cordts2016cityscapes}, which are suitable for the traffic scenes in our dataset.

\Fref{fig:scene_stats}a shows that nearly 100\% of the images contain sky, building, vegetation, road, and car, which is not surprising. In this dataset, bicycle, motorcycle, and rider are also seen very often. 
\Fref{fig:scene_stats}b shows the percentages of scene-image pixels belonging to each semantic class. The majority of the pixels belong to road, sky, building, and vegetation, while the minority ones belong to traffic sign and traffic light with 0.1\% and 0.04\%, respectively. Compared to other classes, traffic lights and signs have smaller sizes and appear less often.



\Fref{fig:gaze_scene_perc} shows the class distributions for the semantics of gaze pixels and all pixels. For each semantic class, the blue bar shows the percentage of times a fixation point belongs to the class, and the red bar is the percentage of pixels in the scene camera belonging to this class. While the percentage of road and car pixels is small, these classes attract the driver's attention the most. On the contrary, the percentages of hood, sky, vegetation, and building pixels are high, but they do not attract the driver's attention. This also shows the driver's mentality of always paying attention to the objects that affect the driving safety such as road, car, and rider. The percentage of the time the driver is looking at a traffic light or traffic sign is about the same with the percentage of their pixels.


\begin{table}[t]
    \caption{\bf Statistics of the proposed DPoG dataset  \label{tab:data_stats}}    
    \setlength{\tabcolsep}{4pt}
    \centering
    \begin{tabular}{lrrr}
    \toprule
    Number of   & Train & Test & Total \\
    \midrule 
    Sessions & 11 & 8 & 19\\ 
    Clip triplets & 354 & 235 & 589 \\
    Frame triplets & 105,951 & 70,500 & 176,451\\ 
    Gaze frames with gaze point & 90,614 & 62,180 & 152,794\\
    Scene frames with gaze point & 85,573 & 58,102 & 143,675\\
    \bottomrule 
    \end{tabular}
\vspace{-5mm}
    \end{table}
    


\section{Drivers' Points-of-Gaze Estimation Network}


Our goal is to learn a model to localize the point of gaze of a driver at every time step.  This task is similar to the screen-based eye-tracking task, except that we have a ``3D display'' instead of a 2D screen. In screen-based eye tracking, the position and pose of the face camera with respect to the display screen is fixed and the distance from the viewer to the point of gaze varies in a small range. Our task, however, is much harder because we need to predict gaze for dynamic environments with unknown varying depths. To tackle this problem, we need to analyze both the face image and the scene image, unlike the screen-based eye tracker that only needs to analyze the face image. In this section, we will describe the proposed \textbf{D}rivers' \textbf{P}oints-of-gaze \textbf{E}stimation \textbf{N}etwork (DPEN), a novel convolutional network that inputs both the face image and the scene for point of gaze estimation. 




\subsection{Network Architecture and Processing Pipeline}

\begin{figure*}[t]
\begin{center}
\includegraphics[width=0.95\textwidth]{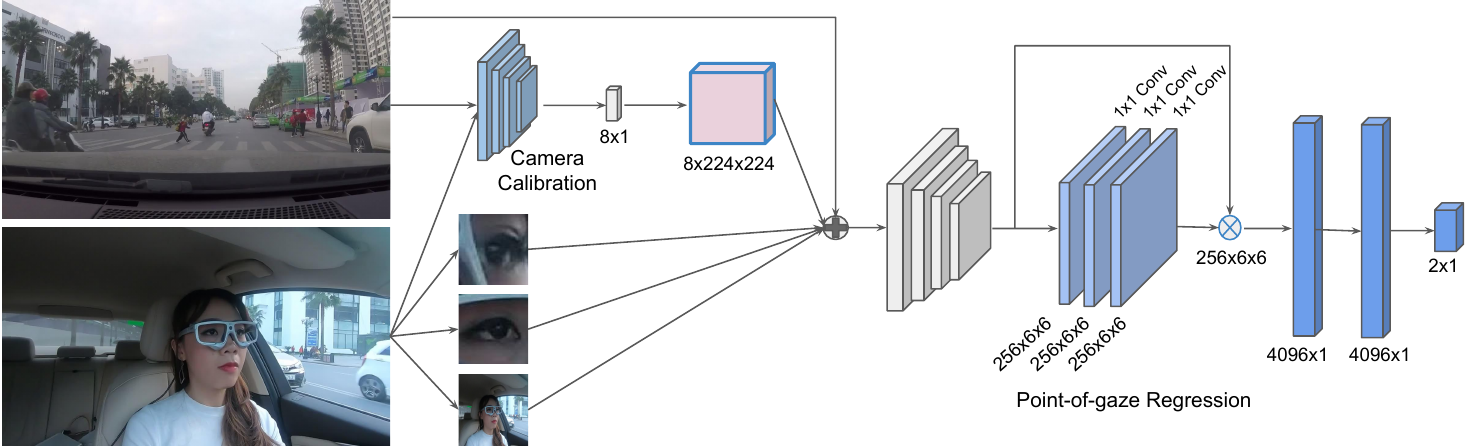}
   \caption{{\bf  Architecture of the proposed Drivers' Points-of-gaze Estimation Network (DPEN). } 
   \label{fig:network_architecture} }
\vspace{-5mm}
\end{center}
\end{figure*}

\Fref{fig:network_architecture} depicts the network's architecture, which has two main components: the camera calibration module and the gaze regression module. The inputs to the network are the face and scene images, and the output is the predicted 2D location for the point of gaze on the scene image. From the face image, we extract a smaller region of interest (ROI) around the face, referred to as the facial ROI. The facial ROI is a fixed window for all images for all driving sessions, which is taken as the smallest window that contains all the faces of the drivers in the dataset. The scene image and the face image will be fed into the camera calibration module to compute an embedding vector for the spatial configuration between the driver and the camera system. The output of the camera calibration module, together with the scene image, the facial ROI, and the eye ROIs are fed into a gaze regression module for the final point-of-gaze estimation. 




\myheading{The camera calibration module} is a ResNet-18~\cite{He-et-al-CVPR16} with an additional residual layer and an average pooling layer. The input to this module is a $6{\times}224{\times}224$ tensor  that is obtained by concatenating the resized scene and face images. The camera calibration module's output is a vector of eight parameters, resembling the set of parameters that relate the position and pose of the two cameras together having a coordinate system centered at the driver's location.



\myheading{The point-of-gaze regression module} consists of a ResNet-18~\cite{He-et-al-CVPR16} followed by a spatial weighting component consisting of three $1{\times}1$ convolutional layers with ReLU activation in between. The output of the spatial weighting component will be multiplied element-wise with the output of the ResNet-18. It then goes through a ReLU and a dropout layer with a dropout rate of 0.5. Finally, there will be two fully connected layers with 4096 dimensions right before the final output of two dimensions.

The inputs to the gaze regression module are the scene image, the facial ROI, the left-eye ROI, the right-eye ROI, and the eight camera calibration parameters. All images are resized to $224{\times}224$. We repeat the camera calibration parameters over spatial dimensions to form an eight-channel map of size $8{\times}224{\times}224$. It is then stacked with other images by channels, forming a 20-channel input.

The eye ROIs are extracted as follows. We first detect the 98 facial landmarks \cite{h3r}. For each eye, its center is taken as the midpoint of the two eye corners. The eye ROI is a squared window centered at the eye center, with the width and height equal to 1.5 the distance between the eye corners.

\subsection{Training procedure}

The proposed network can be trained end-to-end, jointly optimizing the parameters of the point-of-gaze regression module and the camera-calibration module. The main training objective is to minimize the discrepancy between the predicted and the ground truth gaze points, and we optimize a loss based on weighted Euclidean distances. We also add a triplet loss for the camera calibration module. 

Let $\{x_i, p_i\}_{i=1}^{n}$ denote the set of training data, where $x_i$ is a pair of scene and face images and $p_i$ is the annotated point of gaze on the scene image. During training, we minimize the following loss function: 
\begin{align}
    \mL = \frac{1}{n}\sum_{i=1}^{n} \left(\mL_{dist}(x_i) + \mL_{trip}(x_i) \right).
\end{align}
These loss functions are described in details below.

\myheading{Weighted Euclidean distance loss}.  Let $\hat{p}_i$ be the prediction output of the network, we define the main prediction loss  $\mL_{dist}(x_i)$ based on a piece-wise linear function of the Euclidean distance between the annotated and predicted gaze points: $||p_i - \hat{p}_i||_2$ as follows:
\begin{align}
\scalebox{0.9}{
$\mL_{dist}(x_i) = \beta {.} Relu(\|p_i - \hat{p}_i\|_2 - \tau) - \alpha {.} Relu (\tau - \|p_i - \hat{p}_i\|_2)$, \nonumber 
}
\end{align}
This loss is a continuous piece-wise linear function of the distance. If the distance is greater or equal to a threshold~$\tau$, the loss becomes $\beta\|P - \hat{P}\|_2^2 - \beta\tau$. If it is smaller than $\tau$, the loss is $\alpha \|P - \hat{P}\|_2^2 - \alpha\tau$. Here $\alpha, \beta, \tau$ are tunable hyper-parameters. In our implementation, $\alpha = 0.1, \beta = 2, \tau = 5$. We use $\alpha < \beta$ to scale down the loss (and gradient) when the predicted gaze point is already sufficiently close to the ground truth, accounting for the inaccuracy of the automatically derived ground truth gaze points.

\myheading{Triplet loss}. The triplet loss is defined for the camera calibration module. Let $f$ denote the calibration module and let $f(x_i) \in \mathbb{R}^{8}$ be the output embedding vector of the eight calibration parameters. We also constraint the embedding vector to have unit norm: $\|f(x_i)\|_2 = 1$. Desirably, the distance between two embedding vectors for training data instances from the same driving session should be smaller than the distance between two embedding vectors of data instances from different sessions. Thus, for each training instance $x_i$, we randomly sample a training instance $x_i^p$ from the same driving session and another training instance $x_i^n$ from another driving session. $x_i, x_i^p, x_i^n$ are referred as the \emph{anchor}, \emph{positive}, and \emph{negative} data points respectively. The embedding distance between the anchor point and the negative point should exceed the embedding distance between the anchor point and the positive point by a margin $\mu$, i.e.,
\begin{align}
     \left\| f\left ( x_{i} \right ) - f\left ( x^{n}_{i} \right ) \right \|_{2} -  \left \| f\left ( x_{i} \right ) - f\left ( x^{p}_{i} \right ) \right \|_{2} > \mu.  
\end{align}
We therefore define the loss based on this margin violation: 
\begin{align}
\scalebox{0.88}{
    $\mL_{trip}(x_i) = Relu(\left \| f\left ( x_{i} \right ) - f\left ( x^{p}_{i} \right ) \right \|_{2} - \left\| f\left ( x_{i} \right ) - f\left ( x^{n}_{i} \right ) \right \|_{2}  + \mu).$
    } \nonumber 
\end{align}
The margin hyper-parameter $\mu$ is set to $0.2$ in our tests. 

\myheading{Optimization details.} We use Adam optimizer with learning rate of $10^{-4}$ and batch size $256$. In each iteration, for each training data point in the batch of data, we sample the positive and negative data points randomly from the same batch. We run the optimization for 150 epochs, which takes around one day on single NVIDIA V100 GPU.

\section{Experiments}


In this section, we report the comparison of our method its with several baselines and also describe our ablation studies. We use the Euclidean distance between the predicted gaze point and the annotated gaze point as the main performance metric. Additionally, we also use the Area Under the ROC Curve (AUC), a commonly used metric for evaluating saliency prediction models \cite{Bylinskii_etal_2019}. 

\myheading{Image-independent baseline methods.} One simple baseline is to always use center of the scene image as the prediction output. This baseline is motivated by the center bias phenomenon in saliency prediction. Another related baseline is to always predict a fixed position, which is determined as the mean of the fixation points in the training data. This baseline also assumes there is a bias point for the fixation location, and that bias can be estimated using training data. These two baselines are referred to as \emph{Center-point prediction} and \emph{Fixed-point prediction} respectively. 

\myheading{Face-independent baseline methods.} We consider two baseline methods that make prediction based on the scene but not the face image. Particularly, we use an object detector \cite{Tan_2020_CVPR} to detect cars in the scene image and localize the car instance that is closest to the center of the scene image. The predicted gaze point is then taken as the center of the detected car. We refer to this method as \emph{Car-in-front prediction}. This baseline method is motivated by the statistics shown in \Sref{sec:scene_gaze_stats} that the drivers spend a significant percentage of time looking at cars. We also use TASED-Net~\cite{min2019tased}, a state-of-the-art saliency detection network as another face-independent  method. 

\myheading{End-to-end trainable method.} We also consider VideoAttentionTarget~\cite{chong2020detecting}, a state-of-the-art network for predicting where in an image a person in the image looks at. This network was developed for a different task, and it does not have a camera calibration module. We use the authors' implementation and train the network using our training data.

GazeRefineNet \cite{Park2020ECCV} is another state-of-the-art gaze estimation baseline. It uses the left and right eye images of a person in combination with the corresponding screen content to improve the point-of-gaze estimate. Making GazeRefineNet works on our proposed dataset is tricky. It requires the camera's extrinsic parameters as input, which are not available in our dataset. Hence, we have to assume these parameters are fixed across videos, compute their optimal value, and use them for training. We use the static version of GazeRefineNet, a similar configuration as DPEN, for a fair comparison. In their paper, GazeRefineNet used an offset augmentation to adapt gaze estimation for different people. This technique requires an explicit analysis on a video sequence of the target subject at inference time, which is impractical. Hence, we skip that augmentation and use the standard static version. \Tref{tab:rslt} reports GazeRefineNet's error on our dataset as $70.24$, which is much larger than ours.

\myheading{Result discussion.} \Tref{tab:rslt} shows the prediction errors of the proposed method DPEN and all aforementioned baselines. The average prediction error of DPEN is 29.69 pixels, which is relatively small compared to the size $1280{\times}720$ pixels of the scene images. DPEN outperforms other methods by a large margin. Note that VideoAttentionTarget and TASED are the two methods that output a probability map instead of a 2D location. To convert the probability distribution into a point estimate, we use either the Mean or Mode of the distribution. We experiment with both as shown in \Tref{tab:rslt}. We can also compare DPEN with these methods in terms of AUC. The AUC of VideoAttentionTarget, TASED, and DPEN are 0.8545, 0.8793, and 0.9698, respectively. For AUC, the higher, the better. \Fref{fig:error_dist} shows the distribution of the errors and \Fref{fig:qualitative} shows some qualitative results. 

In addition to using pixel for measuring prediction errors, we find that eye-angle would be a more meaningful metric, but it is also harder to measure the eye angle directly due to the unknown scene depths. However, because the distance between the driver's eyes to the scene camera is much smaller than the distance between the eyes to the objects on the street scene, we can approximate the eye angle by the scene camera angle. Based on the field-of-view (FOV) of GOPRO cameras, we convert the prediction errors in pixels to angles and report them in \Tref{tab:rslt}. As can be seen, the eye-angle error of the proposed method DPEN is less than 3 degree, which is sufficiently good for various practical applications.

We also perform an ablation study where we train a DPEN model without the triplet loss. This model has the prediction error of 121.19 pixels and 12.02 degree, which is significantly worse. This clearly demonstrates the benefits of having the triplet loss and also the camera calibration module. We also consider a DPEN model in which the scene image is not fed into the point-of-gaze regression module. One the one hand, this model does not work as well as the one that uses the scene image, demonstrating the benefits of using the scene image for prediction. On the other hand, the increase in the prediction error is not enormous. This is because the point of gaze depends on both the eye direction and the depth of the scene, and perhaps the eye direction is more important than the scene depth and we can obtain a reasonable estimate even without accurate depth estimation. 


\begin{table}[t]
    \setlength{\tabcolsep}{1pt}
    \caption{{\bf Comparing the prediction errors of DPEN and various methods}. DPEN yields the best performance, and both the triplet loss and scene image are crucial for its superior performance  \label{tab:rslt}}      

    \vskip 0.1in
    \centering
    \begin{tabular}{lrc}
    \toprule
    Method &   Error $\downarrow$ & Eye-angle ($^\circ$) $\downarrow$\\
    \midrule
Center-point prediction &    159.95 & 15.87\\
Fixed-point prediction &   140.78  & 13.97\\
Linear Regression       &  124.00 & 12.30\\
Car-in-front Prediction     & 151.82 & 15.06\\
VideoAttentionTarget \cite{chong2020detecting} - Mean & 155.31  & 15.41\\
VideoAttentionTarget \cite{chong2020detecting} - Mode & 198.44  & 19.69\\
TASED-Net \cite{min2019tased} - Mean &   154.53   & 15.33\\
TASED-Net \cite{min2019tased} - Mode &   184.84  & 18.34\\
GazeRefineNet \cite{Park2020ECCV} & 70.24  & ~6.97\\
DPEN (proposed)         &  {\bf 29.69}  & {\bf ~2.95}\\
\midrule 
DPEN without triplet loss  & 121.19 & 12.02\\
DPEN without scene image     &   48.59 & ~4.82\\
    \bottomrule 
    \end{tabular}
    \vskip 0.1in
    \end{table}

\def\subFigSzFour{0.48\linewidth}
\begin{figure}[t]
\centering 
\includegraphics[width=\subFigSzFour]{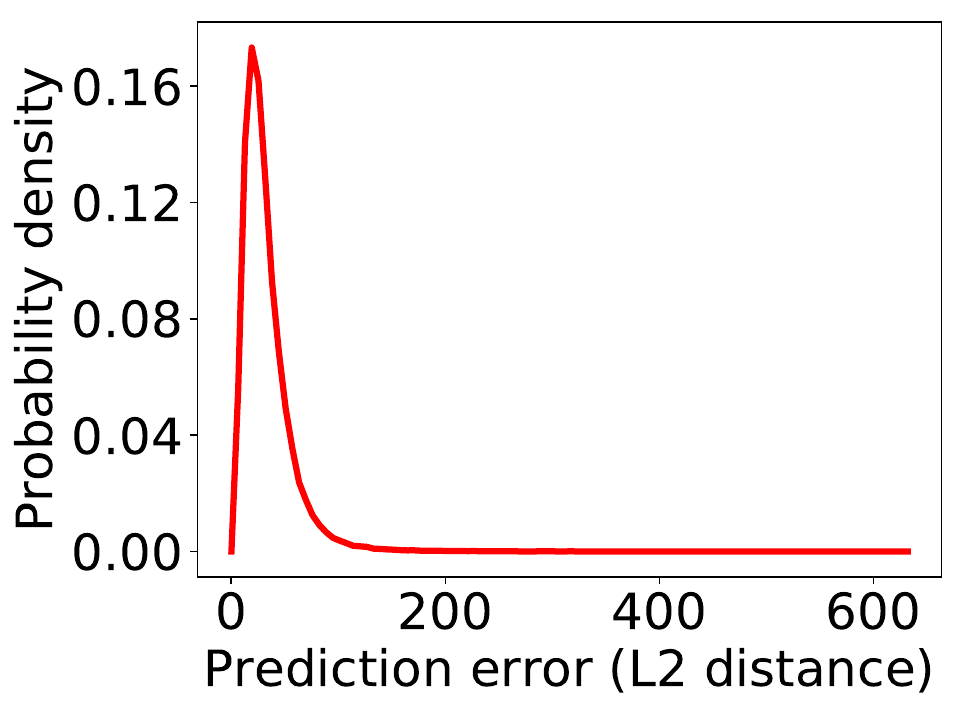}
\hspace{0.25mm}
\includegraphics[width=\subFigSzFour]{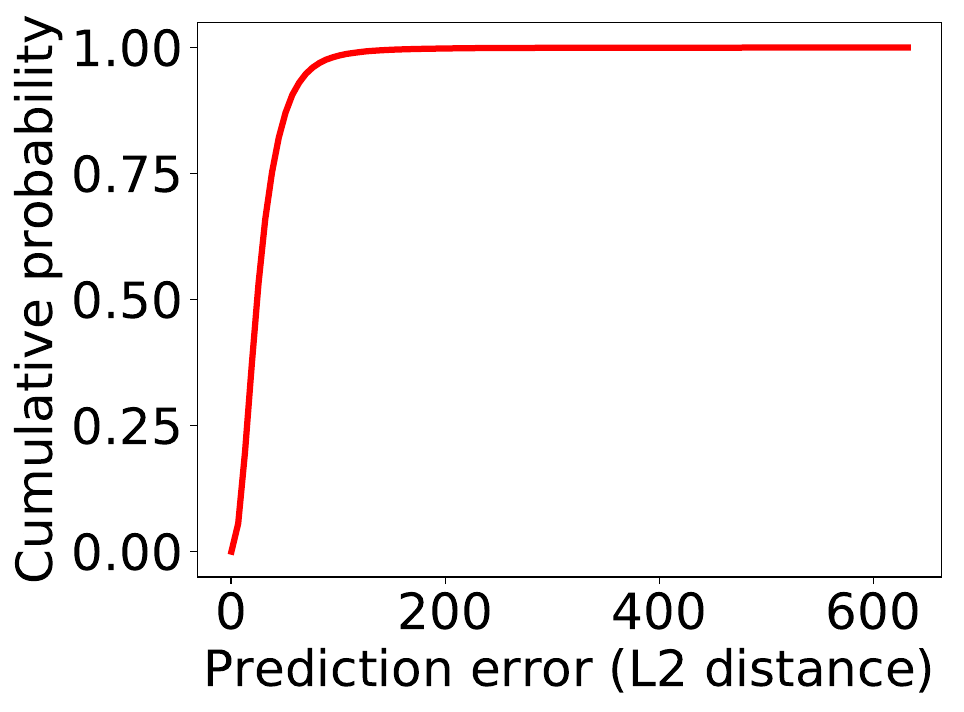} \\
   \caption{{Distribution of the prediction errors.} Left: probability density function, right: cumulative probability. The Mean and Median are: 29.69 and 24.05.} 
\label{fig:error_dist}
\end{figure}

\def\subFigSzThree{0.24\linewidth}
\begin{figure*}[h]
\centering 
\makebox[\subFigSzThree]{\small{Error: 14.60 pixels }}
\makebox[\subFigSzThree]{\small{Error: 19.45 pixels }} 
\makebox[\subFigSzThree]{\small{Error: 42.02 pixels }}
\makebox[\subFigSzThree]{\small{Error: 67.26 pixels }}
\includegraphics[width=\subFigSzThree]{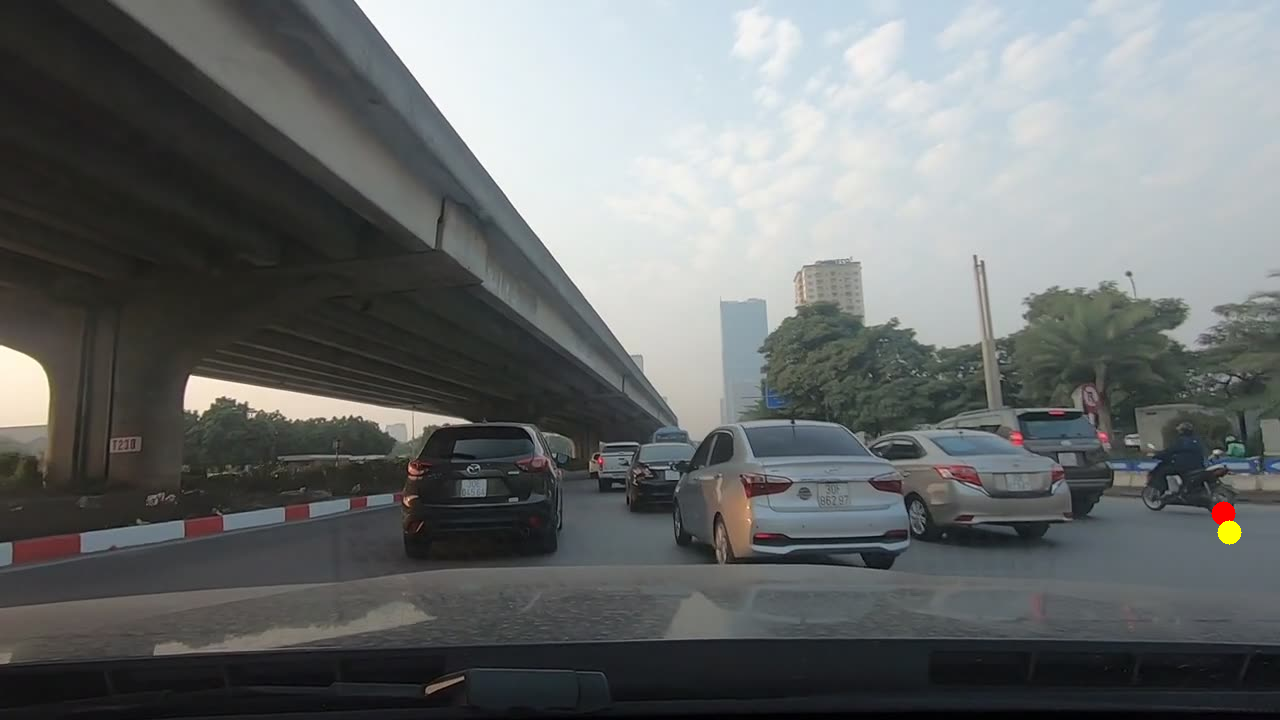}
\includegraphics[width=\subFigSzThree]{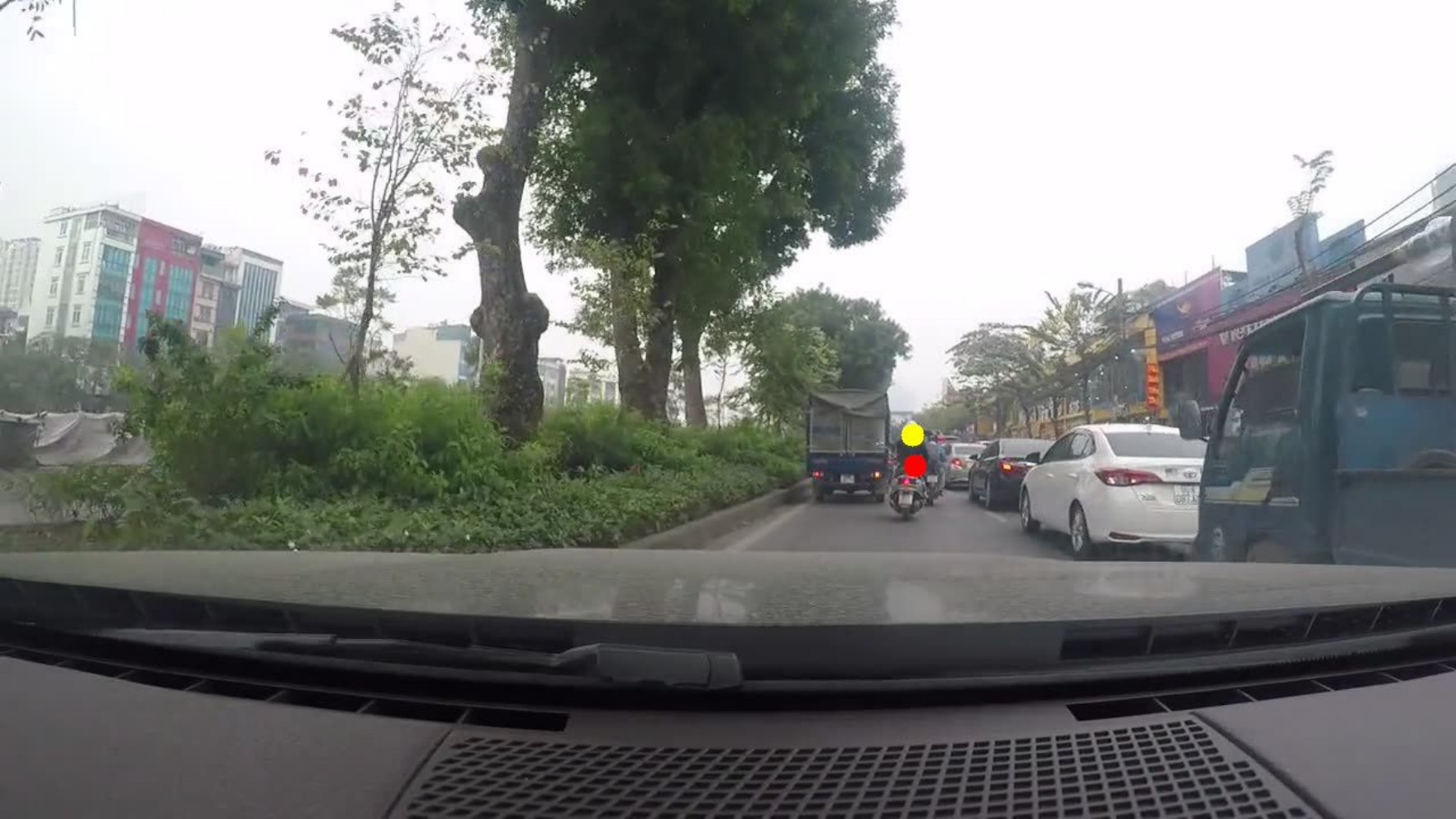} 
\includegraphics[width=\subFigSzThree]{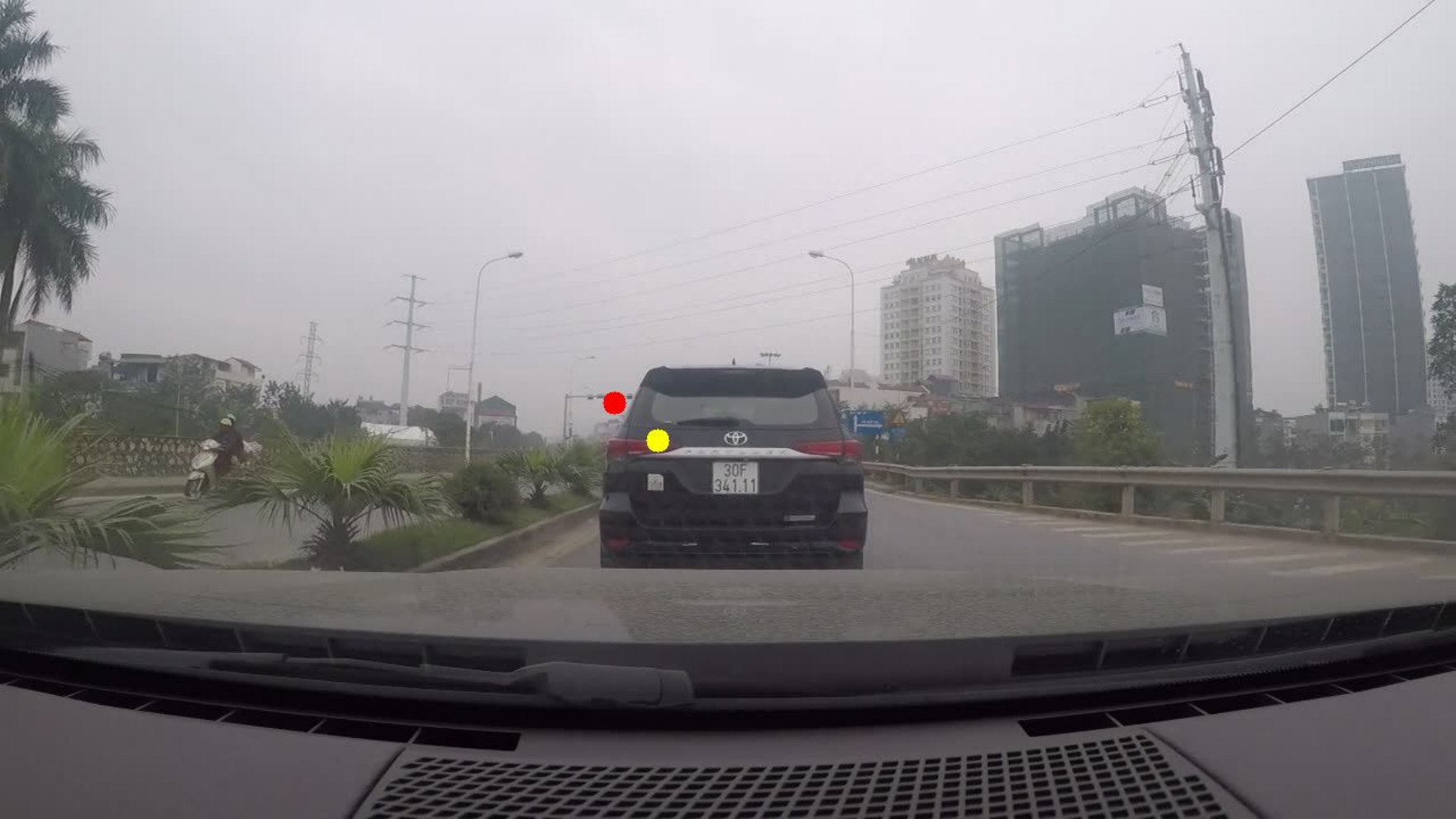}
\includegraphics[width=\subFigSzThree]{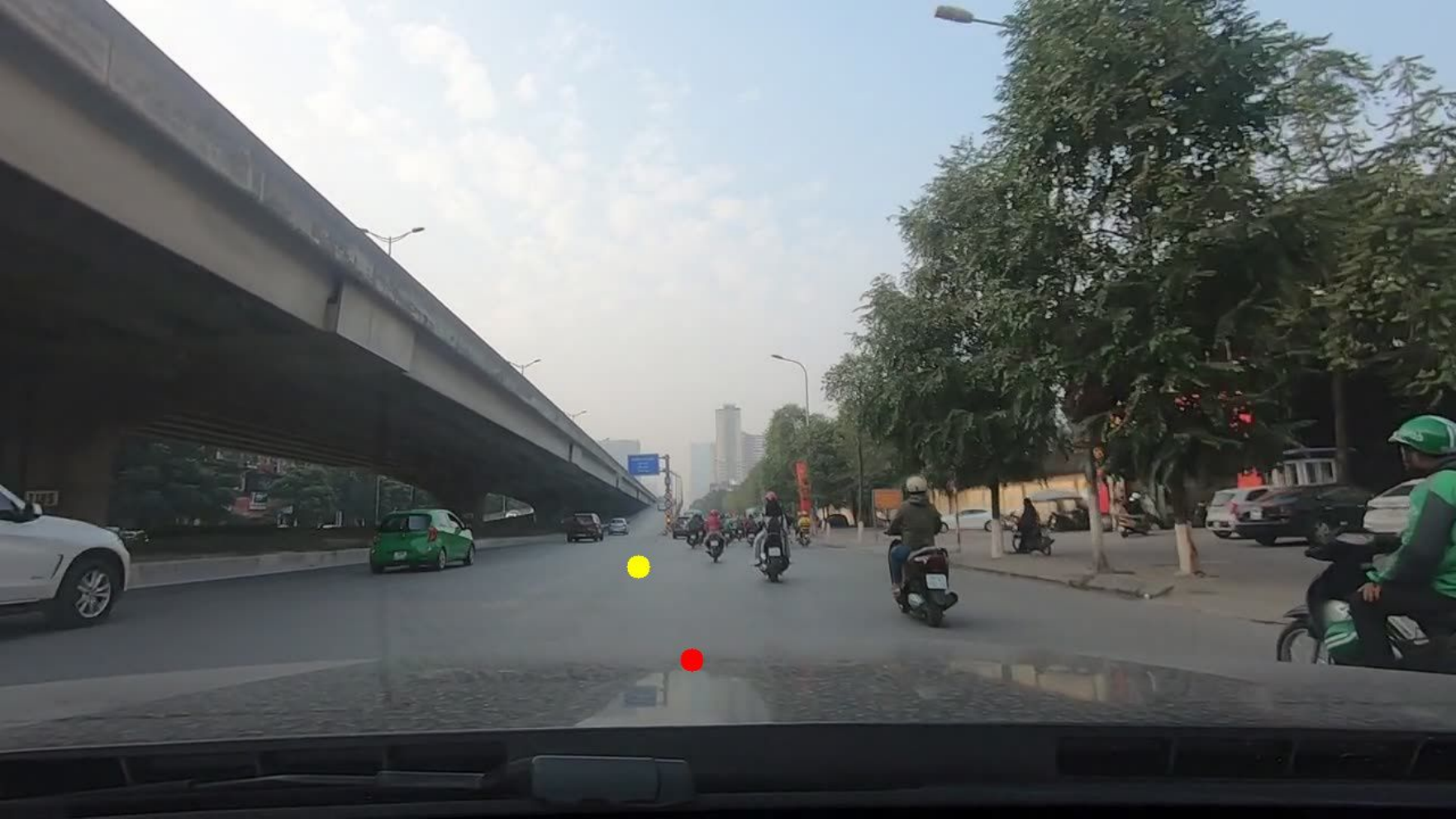}
   \caption{Representative prediction results. The automatically-annotated gaze point is shown in red, while the estimated gaze point by DPEN is yellow.}
\label{fig:qualitative}
\end{figure*}

\begin{figure}[h]
\centering 
\includegraphics[width=\linewidth]{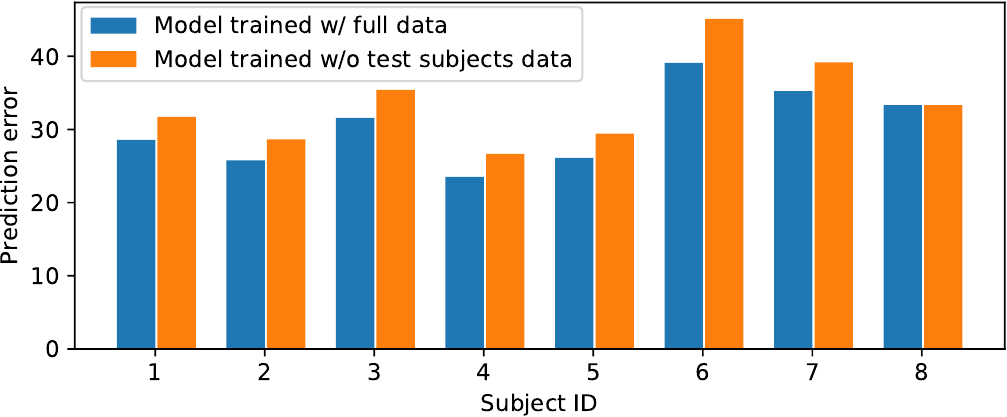}
\vskip 0.1in 
   \caption{Comparing the DPEN model trained with all training data sessions with the leave-one-subject-out DPEN models where the training data session from the test subject is not used for training. The leave-one-subject-out models have similar prediction errors to the full DPEN model.}
\label{fig:full_vs_leave-one-out}
\end{figure}



We also perform a leave-one-subject-out experiment. Recall that the test data contains eight driving sessions from eight subjects. For each testing session, we identify the driver and train a DPEN model without using the training data session from this driver, and then test on this testing session. Thus, we have eight leave-one-subject-out models, and \Fref{fig:full_vs_leave-one-out} compares the performance of these models with the model trained with all training data sessions. The leave-one-subject-out models perform well, proving the generalization ability of our method to new drivers.  

\myheading{Validation on the EVE dataset.} We would like to validate our method on another dataset, but unfortunately no other suitable datasets exist. The closest one we found was the EVE dataset~\cite{Park2020ECCV}, which contained points of gaze on a calibrated 2D computer screen. However, a calibrated 2D screen is very different from an 3D scene with unknown depth. Nevertheless, we modified DPEN to work on this dataset for verification purposes. We did not use the camera calibration module because this was not needed for the 2D aligned screen. We extract $147,470$ frames from EVE video for training DPEN and evaluate it on $17,350$ other frames. We compare with the static version of GazeRefineNet without offset augmentation as described earlier. The prediction errors for GazeRefineNet and DPEN are 127.59 and 128.01, respectively. DPEN is slightly worse than GazeRefineNet, but GazeRefineNet has many unfair advantages over DPEN because it was specifically developed for 2D screen explicitly taking into the known calibration parameters.

\section{Conclusions}


We have presented a dashboard-mounted eye-tracking system for tracking a drivers' points of gaze in traffic environments. This system consists of two cameras, looking at the driver's face and the road. To accompany this eye-tracking system, we have developed a method that can  account for the volatile distance between the driver and the camera system to estimate the driver's point of gaze. This method achieves relatively low mean prediction error with respect to the resolutions of the scene camera. We have also introduced a large-scale dataset with ground truth and automatically transferred points-of-gaze annotation. This is a rich dataset, and we have reported various interesting scene and behavioral statistics for the real driving sessions. 


{\small
\bibliographystyle{ieee_fullname}
\bibliography{pubs_egpaper}
}

\end{document}